\documentclass[11pt,a4paper]{article}

\usepackage{amsmath,graphicx,fullpage}
\usepackage{cite}
\usepackage{booktabs}
\usepackage{multirow}
\usepackage{array}
\usepackage{enumerate}
\usepackage{color}
\usepackage{url}
\usepackage{pifont}
\newcommand{\cmark}{\ding{51}}%
\newcommand{\xmark}{\ding{55}}%
\usepackage{pdflscape}
\newcolumntype{x}[1]{%
>{\centering\hspace{0pt}}p{#1}}%

\usepackage{fancyhdr}
\date{}
\title{\Large {\bf BVI-Lowlight: Fully Registered Benchmark Dataset for Low-Light Video Enhancement}}
\author{Nantheera~Anantrasirichai, Ruirui~Lin, Alexandra~Malyugina, and~David~Bull\footnote{This work was supported by the UKRI MyWorld Strength in Places Programme (SIPF00006/1) and Bristol+Bath Creative R+D under AHRC
grant (AH/S002936/1)} \\  Visual Information Laboratory \\ University of Bristol}

\begin{document}
\maketitle

\begin{abstract}
Low-light videos often exhibit spatiotemporal incoherent noise, leading to poor visibility and compromised performance across various computer vision applications. One significant challenge in enhancing such content using modern technologies is the scarcity of training data.
This paper introduces a novel low-light video dataset, consisting of 40 scenes captured in various motion scenarios under two distinct low-lighting conditions, incorporating genuine noise and temporal artifacts. We provide fully registered ground truth data captured  in normal light using a programmable motorized dolly, and subsequently, refine them via image-based post-processing to ensure the pixel-wise alignment of frames in different light levels.  This paper also presents an exhaustive analysis of the low-light dataset, and demonstrates the extensive and representative nature of our dataset in the context of supervised learning. Our experimental results demonstrate the significance of fully registered video pairs in the development of low-light video enhancement methods and the need for comprehensive evaluation. Our dataset is available at \url{DOI:10.21227/mzny-8c77}.

\end{abstract}

\section{Introduction}

Low-light video footage is problematic due to the interactions between the aperture, shutter speed and ISO parameters which cause distortions, particularly under extreme lighting conditions.  This is due to the inverse correlation of increased photon noise with decreasing light intensity where increased sensor gain amplifies noise. Secondary characteristics such as white balance and color effects will also be affected and will require correction.  While these distortions cause a reduction in perceived visual quality,  the associated artefacts also pose significant challenges to machine learning tasks such as classification and detection. This is especially the case when one considers the effect of adversarial examples on some deep learning networks.

\begin{figure}[t]
  \centering
    \includegraphics[width=0.7\linewidth]{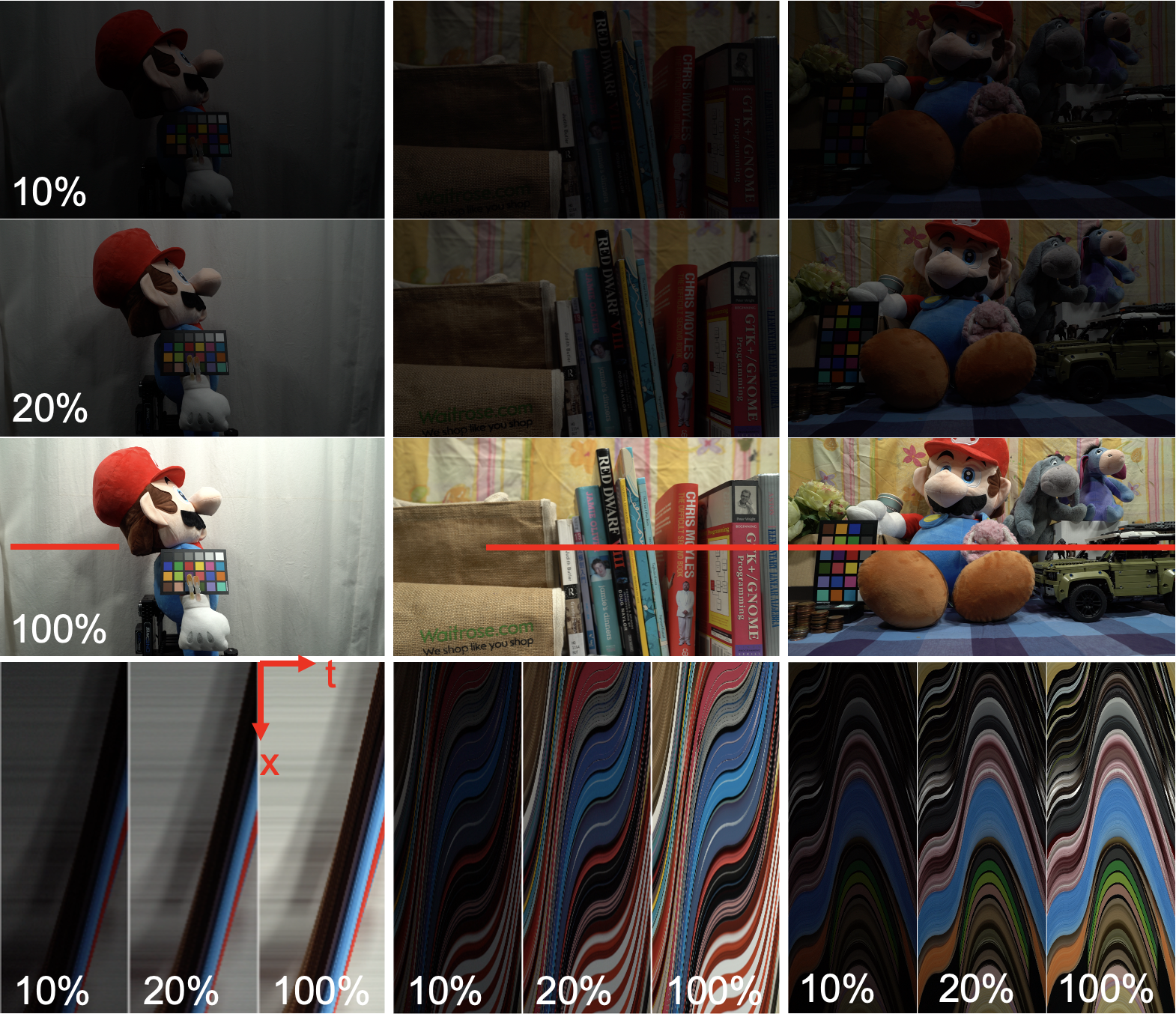}
  \caption{Scene examples with varying light levels (10\% and 20\% low-light, 100\% normal light) and different motion profiles (bottom row). From left to right columns: Mario3 with a static background, Book1, and Mario1. The length of the red lines across the normal-light frames represents the height of the x-t planes in the bottom row.}
  \label{fig:motions}
\end{figure}

The performance of low-light \textit{image} enhancement algorithms has been significantly improved in recent year, primarily through the application of deep learning methods. Such techniques, however, cannot be applied directly to low-light \textit{video} content as their application on a frame-by-frame basis causes temporal inconsistencies. Optimization of these methods for motion video is, in turn, limited by the need for large and diverse datasets with aligned ground truth information. However, low-light video enhancement is an ill-posed problem, meaning that clean ground truth information with accurate brightness is not available. 

Improving video footage shot under low-light conditions is highly required and is emerging as an important research field, particularly within the context of deep learning methods.  In general these solutions, especially supervised learning architectures, require  aligned and accurate ground-truth data to support robust training. This however is limited for low-light video research. To address this, researchers frequently resort to the use of synthesized noise.  Self-supervised learning methods, which do not rely on ground truth data, can be applied to denoising tasks \cite{Dewil:Self:2021}. This is possible because the details are randomly collapsed, and content priors are inhibited in the videos. However, while these methods can achieve contrast enhancement to a certain extent, akin to applying histogram stretching, they are ill-suited for color correction tasks. The reason is that low-light images and videos lack accurate color information, rendering color correction impossible. This has been confirmed by empirical tests reported in \cite{Liang:self:2022}. Alternatively, a pseudo reference with a unpaired learning strategy can be used \cite{anantrasirichai:Contextual:2021}, but the methods generally are often complex and require large memory.

This paper presents a dataset of fully registered sequences between low-light videos and their ground truth, shown in Fig. \ref{fig:motions}. It was captured over three varying lighting conditions, including the 100\% lighting being classed as ground-truth. The dynamic scenes were created with programmable motorized dolly, allowing repeatable motion profiles. Our dataset comprises a diverse range of forty scenes, containing numerous contents, varying structures and textures. This results in more than 30k ground-truth paired frames. It is worth noting that normal light levels can be conveniently adjusted to achieve the desired exposure, thereby creating pseudo ground-truth data to evoke different cinematic moods and tones. Additionally, we chose to capture the videos in the standard RGB format, whereas many datasets are in the raw Bayer format. This decision makes our dataset more accessible to the research community since consumer cameras typically do not provide raw videos.

We also present an exhaustive analysis of the low-light dataset, focusing on noise, color correction, contrast enhancement in both spatial and temporal information. Moreover, we demonstrate the extensive and representative nature of our dataset in the context of supervised learning by employing it to train state-of-the-art models for image and video enhancement. Our experimental results show that the significance of fully registered video pairs in the development of low-light video enhancement methods and comprehensive evaluation.

In conclusion, our main contributions are as follows:
\begin{itemize}
    \item We present the high-quality video dataset captured in motion under three distinct real lighting conditions, offering 40 different dynamic scenes with diverse content.
    \item Our dataset is the fully registered videos (spatially aligned frames between different lighting videos) in each scene, facilitating supervised learning and enabling full-reference quality assessment.
    \item Our dataset provides two specific light levels, rather than random variations in brightness. This design allows for a more in-depth analysis of the effects of different light levels, contributing to a better understanding of their characteristics and influence on deep learning architectures.
    \item We conducted extensive experiments to demonstrate the impact of fully-registered videos, diverse content, and motion on the performance of low-light video enhancement methods.
\end{itemize}

\section{Existing low-light video datasets}
\label{sec:existdataset}

This section discusses datasets available for low-light video enhancement. While the number of datasets captured under real low-light conditions is limited, synthetic samples have expanded research activities in this field. However, it is important to note that synthetic distortions are often oversimplified, as mentioned earlier, low-light conditions combine multiple distortion types.

\subsection{Real datasets}
\begin{figure}[t]
    \centering
    \includegraphics[width=0.7\columnwidth]{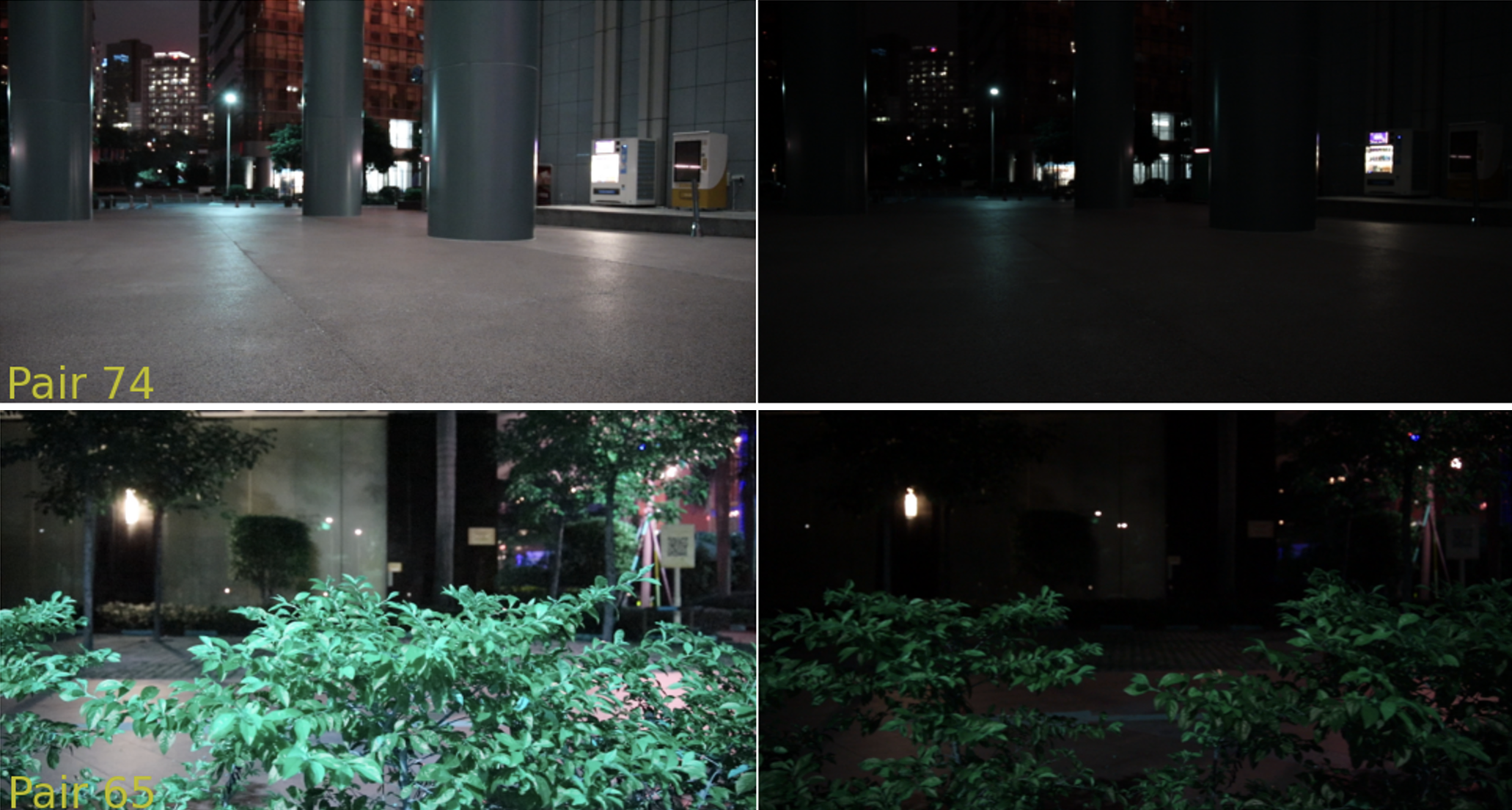}
    \caption{Examples of faulty SDSD video pairs, where the ground truth and the low-light frames are on the left and the right, respectively. (Top) The overexposed areas in the ground truth frame. (Bottom) The unaligned pair.}
    \label{fig:SDSD_samples}
\end{figure}

Several `\textit{image}'-based datasets exist that can be used to support denoising and other inverse problems. These include the UCID-v2 corpus \cite{5651245}, the Darmstadt Noise Dataset \cite{8099777}, the Smartphone Image Denoising Database (SIDD)~\cite{abdelhamed2018high}. There are a couple of low-light video datasets captured in static scenes, allowing for the creation of ground truth. This includes BVI-LOWLIGHT, which offers true ISO noise in videos captured at 30 frames per scene \cite{Malyugina:topological:2023}, and DRV (also referred to as SMID) \cite{Chen_2019_ICCV}, capturing 60-100 frames per scene. The DRV also provides dynamic videos without ground truth, similar to the BBD-100K dataset \cite{Yu_2020_CVPR}.  Notably, only three datasets that feature motion and provide ground truth to date:  

i) SMOID \cite{Jiang:learn:2019}: The primary purpose of this dataset is for video surveillance. The dataset consists of 179 pairs of street view videos containing moving vehicles and pedestrians. These video pairs were captured simultaneously using a beam splitter, and an ND (Neutral-density) filter was inserted to limit the amount of light in one camera to simulate low-light environment. Only the overlapping area was used, resulting in a frame resolution of 1800$\times$1000 pixels. Each video has 200 frames with a GBRB Bayer pattern. Unfortunately, this dataset is not publicly available.

ii) RViDeNet \cite{9156826}: It has been introduced specifically for video denoising purposes, where the videos were captured at 5 different ISO levels, ranging from 1600 to 25600. Each video contains only 7 frames with a resolution of 1920$\times$1080 pixels in Bayer format. They were captured using the stop motion technique, so there is no motion blur as in reality. Also, the displacement between frames is very large resulting unnatural footage. The corresponding clean frame is obtained by averaging multiple noisy frames, resulting in a clean frame that is still dark. This means that this dataset cannot be used for brightness enhancement and color correction, as required for low-light video enhancement.
  
iii) SDSD \cite{wang2021sdsd}: The video pairs were captured using a motorised dolly. However, their video frames are not correctly aligned, making them unsuitable for supervised learning or objective evaluation. Our testing experiments have confirmed this issue. The dataset comprises 150 video pairs featuring both indoor and outdoor scenes, each consisting of 100-300 frames per video and having a resolution of 1920$\times$1080 pixels. Although they claimed that their darker videos were created with an ND filter and adjusted ISO settings, we found inconsistencies in the lighting conditions, as some video pairs exhibit different shadows.  Unfortunately, there are no specific details available regarding the camera settings they used. Moreover, several reference videos are overexposed, making it impossible to recover the details in these areas.  This is confirmed by their analysis, which reveals that some low-light patches have higher intensity than their corresponding normal-light patches. Examples of the SDSD datasets are shown in Fig. \ref{fig:SDSD_samples}.

iii) DID \cite{Fu:dancing:2023}: There are 413 pairs of videos, each with a resolution of 2560$\times$1440, captured with five different cameras. The dynamic scenes were created using an electric gimbal that rotated around the camera axis. This setup however results in only one type of motion--panning. The video brightness was adjusted using an ND filter. 

Table \ref{tab:existingdata} shows the existing low-light datasets. These current datasets are obviously limited to develop the reliable tools for video enhancement, either because they are too small or because they lack realistic distortions or ground truths for moving objects.  This consequently impacts their suitability for training and validating new networks. Moreover, the low-light condition of most datasets providing ground truth were created with ND filters. While ND filters are intended to reduce the amount of light entering your camera uniformly across all levels of intensity in the scene, they may exhibit variations in spectral transmittance, potentially leading to color shifts. 
Moreover, ND filters can introduce optical aberrations, reduced sharpness, reflections and lense flare, and exhibit potential degradation over time, impacting image quality and reliability in the experiments \cite{aphalo2020neutral}. 

\begin{table}
	\centering
	\caption{Low-light video dataset comparison. S and D indicate static and dynamic scenes, respectively. 1D means one type of movement. D$^*$ means stop motion. $^\dagger$SMOID dataset is not publicly available. }
	\small
		\begin{tabular}{c|ccccc}
		\toprule
          Name &  Light & Motion & Frames & GT & Register\\
	    \midrule
        BVI-LOW \cite{Malyugina:topological:2023} & Real & S & 13,200 & \cmark & \cmark \\
        DRV \cite{Chen_2019_ICCV} & Real & S & 20,847 &  \cmark & \cmark \\
        DRV \cite{Chen_2019_ICCV} & Real & D & 2,397 & \xmark & \xmark \\
        BBD-100K \cite{Yu_2020_CVPR} & Real & D & - & \xmark & \xmark \\
        SMOID$^\dagger$ \cite{Jiang:learn:2019} & ND filter & 1D & 35,800 & \cmark  & \cmark  \\
        RViDeNet \cite{9156826} &  Real & D$^*$ & 385 & \cmark  & \cmark  \\
        SDSD \cite{wang2021sdsd} & ND filter & 1D & 37,500 & \cmark  & Mix \\
        DID \cite{Fu:dancing:2023} & ND filter & 1D & 41,038 &  \cmark &  \cmark\\
        \hline
        Ours & Real  & D & 31,800 & \cmark & \cmark \\
        \bottomrule
		\end{tabular}
	\label{tab:existingdata}
\end{table}

\subsection{Synthetic datasets}
The simplest approach applies Gaussian noise to the clean image as additive noise \cite{Zhang:beyond:2017}. Basic models like these do not effectively capture the noise found in natural images, especially in low-light conditions, where photon noise is dominant \cite{Malyugina:topological:2023}. Authors in \cite{Wang:enhancing:2019} have analyzed their low-light videos captured with Canon 5D mark III and found that dynamic streak noise (DSN) is dominate. They generate the synthetic noise based on DSN, clipping effect,  shot noise and  photoelectrons.  Note that we do not observe DSN in our videos captured with a Sony Alpha 7SII camera or professional cameras, e.g. RED and ARRI used by filmmakers.
SIDGAN \cite{triantafyllidou2020low} has been proposed to map videos found `in the wild' into a low-light domain, producing synthetic training data. This work is based on dual CycleGAN, which map videos captured with short exposure (low light) into the synthetic long exposure domain (normal light). The authors found that training the model with a combination of real and synthetic data achieves the best performance in video enhancement.

\section{State-of-the-art methods for video enhancement}
\label{sec:existmethods}

With recent advances in deep learning technologies, `\textit{image}' enhancement has made significant progress. In contrast, learning-based video enhancement techniques are still in their early stages, primarily due to more unknown parameters, limitations in datasets, computational complexity, and high memory requirements. Applying image-based techniques sequentially to each frame in the video is a common approach but often results in temporal inconsistencies. An in-depth survey of learning-based methods for low-light image and video enhancement proposed between 2017 and 2021 can be found in \cite{Li:Low:2022}, which also shows that there were only a few methods specifically proposed for video enhancement.

\subsection{Low-light image enhancement}
We have reviewed recent low-light image enhancement methods that could potentially be extended to video sequences. In \cite{Xu:SNR-Aware:2022}, the noise map is simply estimated by subtracting the noisy image from the blur-filtered image, and it then guides the attention mechanism in the transformer. In \cite{Ma:toward:2022}, a self-calibrated illumination (SCI) framework has been proposed to learn the residual representation between the illumination and low-light observation. Recently, deep unfolding approaches have garnered significant interest. An adaptive total variation regularization method has been proposed in \cite{Zheng:adaptive:2021}, employing a network to predict noise levels. The technique in \cite{Wu:URetinex:2022} integrates the Retinex theory model, where the image is an element-wise multiplication of reflectance and illumination, with learnable physical priors for reflectance and illumination.

\subsection{Low-light video enhancement}

Multiple input frames are typically used, denoted as $I_{t+i}, i \in [-N, N]$, where $N$ represents the number of neighboring frames at time $t$. In the case of fully registered frames, such as those from static scenes, averaging pixel values across the temporal dimension results in smoothness and noise reduction. However, this simple strategy is challenged by moving objects or camera motion. SMOID \cite{Jiang:learn:2019} modified UNet to handle multiple frames, assuming that the multiple convolutional layers could handle the displacement between frames due to moving objects. However, this method may not effectively handle fast motion. Therefore, several methods include an alignment module, which aligns the feature maps of neighboring frames with those of the current frame \cite{wang2021sdsd, zhou2021rta}. Earlier work exploits LSTM (Long Short-Term Memory) \cite{Wang:enhancing:2019}, whilst later, Deformable Convolution Networks (DCN) \cite{Dai:Deformable:2017} are more commonly employed for this purpose.
The alignment module still sometimes struggles to perform perfect motion compensation, leading to artifacts when combining features from multiple frames. To address this issue, the authors in \cite{zhou2021rta} include iteratively re-weighting features to minimise merging errors.   Authors in \cite{Liu:low:2023} proposed synthetic events from the video to adaptively combine multiple frames according to speeds of moving objects. Attention modules are utilised in fusing synthetic guided event and spatial features. 

Most methods extract features using ResNet, UNet or a combination of them. DRV \cite{Chen_2019_ICCV} has two ResUNet networks that are shared weights, which are trained on the two random frames of static videos, based on an assumption that the model can extend to dynamic videos. Similarly, SIDGAN \cite{triantafyllidou2020low} trains their networks with shared weights. SDSD \cite{wang2021sdsd} separates noise $n_t$ and illumination $S_t$, generating output $\hat{I}_t = (I_t - n_t)/S_t$.
Some methods require Bayer raw video inputs \cite{Chen_2019_ICCV, Jiang:learn:2019}, which are not suitable for many consumer cameras.

As availability of paired datasets is limited, there are also some attempts that use unpaired training strategy, e.g. using CycleGAN \cite{anantrasirichai:Contextual:2021}. This architecture is also employed to map the low-light RAW to RGB format \cite{triantafyllidou2020low}.

\subsection{Video restoration} 

Deep learning-based techniques for video restoration have demonstrated their effectiveness in tasks like denoising, deblurring, and dehazing. Typically, these networks are trained to address the specific type of distortions. However, challenges arise when natural videos, particularly those captured in challenging environments like low light, exhibit a combination of various distortion types. One of the reasons might be due to the lack of reliable datasets. Therefore, we believe that it is valuable to mention and assess the performance of these networks on our datasets. One notable state-of-the-art model is EDVR \cite{Wang:EDVR:2019}, which employs DCN and attention-based modules to integrate spatial and temporal information. Their applications are super-resolution and deblurring. Vision transformer for videos \cite{liang2022vrt} employs temporal mutual self-attention, which is focused on mutual alignment between two neighbouring frames, while self attention is used for feature extraction. Subsequently, the same researchers integrated recurrent feature refinement into the framework, allowing the target frame to benefit from the restoration of previous frames and reducing memory requirements \cite{liang2022rvrt}. The networks have been tested on low-resolution, blurred, and noisy videos separately.

\section{Proposed low-light dataset}
\label{sec:newdataset}

\subsection{Environment settings}

Acquisition of the new dataset was conducted in a dedicated studio, designed to permit full and precise control of the environment's light intensities.  Four Cineo MavX lights, with up to 8,000 lumens output, were configured in Digital Multiplex (DMX) mode and wired to a Zero 88 FLX S24 lighting console.  The color temperature (CCT) was set to 6,500 Kelvin and the range of lighting intensities.  
Each scene contains a `Datacolor SpyderCheckr24: 24 color Patch and Grey Card', typically used for camera calibration and as a white balance target. This provides a set of homogeneous patches as a reference.

\begin{figure}
  \centering
    \includegraphics[width=0.7\linewidth]{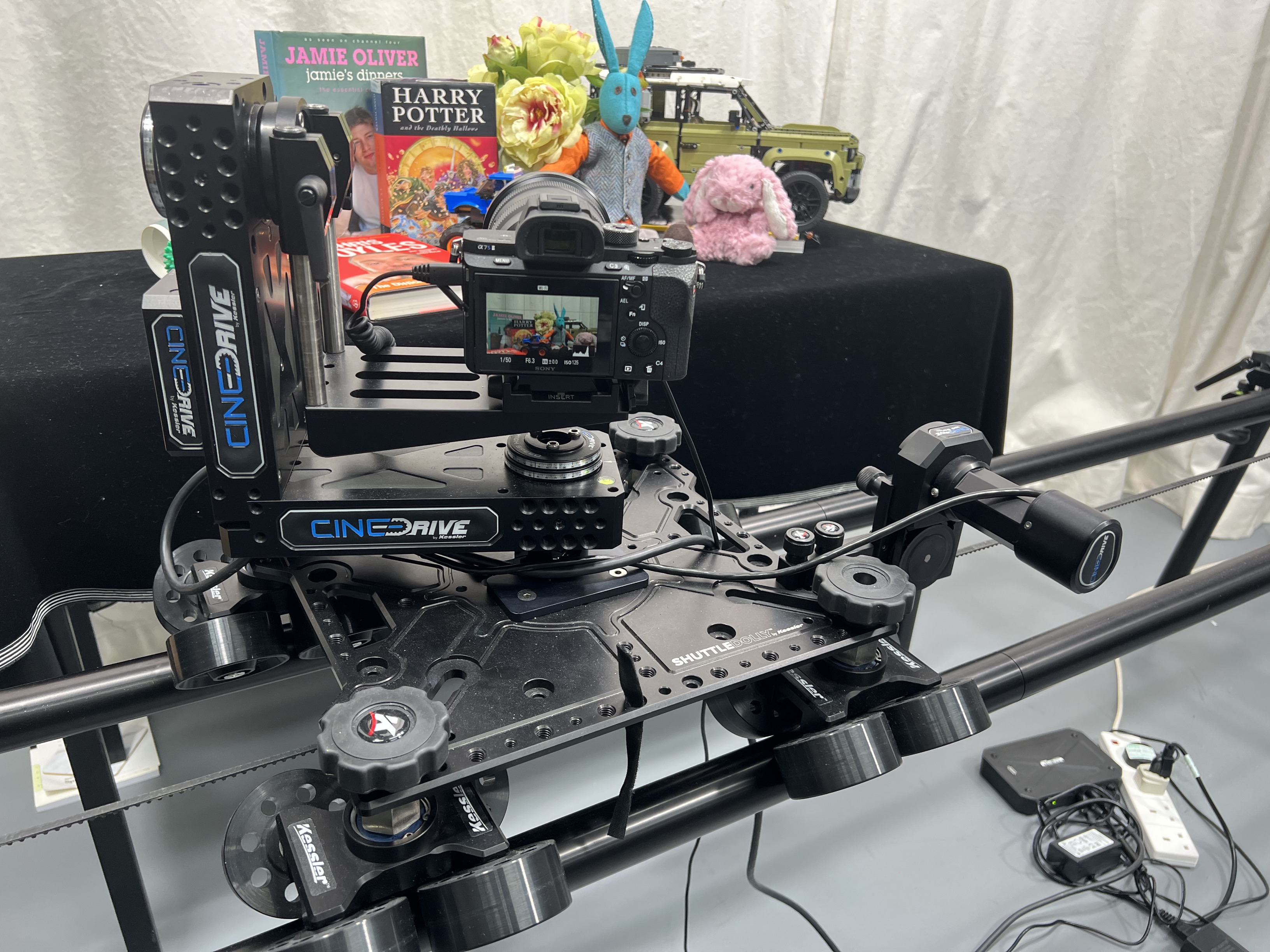}
  \caption{Setting environment showing Sony Alpha 7S II camera in `angle' position, mounted on CineDrive system, with scene in the background.}
  \label{fig:cinedrive}
\end{figure}

We used a Sony Alpha 7SII Wi-Fi enabled camera, capable of recording 4K resolution (3840$\times$2160 pixels) video footage. The repeat movement was controlled using a Kessler CineDrive shuttle dolly system. The CineDrive system, mounted on the dolly allowed additional horizontal camera movement and rotation in three axes, as shown in \ref{fig:cinedrive}. The frame rate was set to 25 fps, which led to a fixed shutter speed of 1/50, as recommended by professionals to achieve a shutter angle of 180 degrees (equivalent to half the frame rate). The aperture was set at F6.8, which fell within the middle range of the camera's optimal aperture setting, producing the highest image quality. In the low-light environments, the ISO was fixed at 800, the optimal value recommended by the manufacturer. It is worth noting that with the Sony camera we used, in-camera noise removal occurs when a higher ISO is employed.

Camera triggering to record a scene is performed using Sony's Imaging Edge mobile phone app, where the phone is connected to the camera via Wi-Fi.  This app also supports the configuration of settings such as aperture, white balance and shutter speed. The use of this remote app avoids unwanted shadows in the footage and sensor misalignments that could arise if the camera was manually triggered.

In this dataset of video pairs, we recorded low-light videos at 10\% and 20\% of normal lighting levels (100\%), as indicated by the Zero 88 FLX S24 light controller. These videos are provided in full HD resolution (1920x1080 pixels), with cropping applied to focus on the region of interest. To ensure that the objects were visible throughout the entire length of the clipped videos, we manually selected the start and stop points.


\subsection{Ground truth generation}
\label{ssec:gtgen}
The primary sources of misalignment are related to the mechanical tolerances of the shuttle dolly and the acceleration of the control motor. Despite careful setup of the mechanical system, there is a possibility of frame misalignment between any two identical motion profiles. We keep the low-light videos as they are because adjusting them could alter the noise characteristics and motion blur. Therefore, we generate pseudo ground truth that is pixel-wise registered to each frame of the low-light videos as follows.

The ground truth videos were recorded under normal lighting condition. To minimize misalignment between the ground truth and the low-light videos, we repeated the dolly movement to create three videos for slow motion and five videos for fast motion. This introduced slight pixel shifts among them, ensuring that the best match to the low-light video had the smallest possible misalignment. In most cases, the best matching video remained consistent throughout the entire video. However, in a few video pairs, the best matching video changed due to a change in the dolly's direction in the middle of the video, such as moving from left to right and then turning back to the left.

The best video matching and frame synchronization were performed by firstly cropping the video by 100 pixels from the borders of the frame to avoid the influence of the background. We subsequently enhanced the frame $t$ of the low-light video $J$ by matching its histogram of each color channel to the histogram derived from the normal-light frame $I^m_n$ of the video $m$ at frame $n$, i.e. $\hat{J}_t = \mathcal{H}(J_t, I^m_n)$, where $\mathcal{H}$ is a histogram matching function with 255 bins for images with an 8-bit depth. This can be seen as contrast enhancement and color correction with respect to the information from the reference (normal light in our case).
Then, the best-aligned reference frame (normal light level) was found by minimising the mean absolute error between $\hat{J}_t$ and $I^m_n$, where $n \in [t-N_w, t+N_w]$, $N_w$ is a predefined searching window range, $m \in [1, N_v]$, and $N_v$ is the total number of videos repeatedly captured at the same normal-light scene. Fig. \cite{fig:bunny} shows error maps depicting the differences between the low-light frame and the normal-light frame before and after frame synchronization.

\begin{figure}[t]
  \centering
    \includegraphics[width=1.0\linewidth]{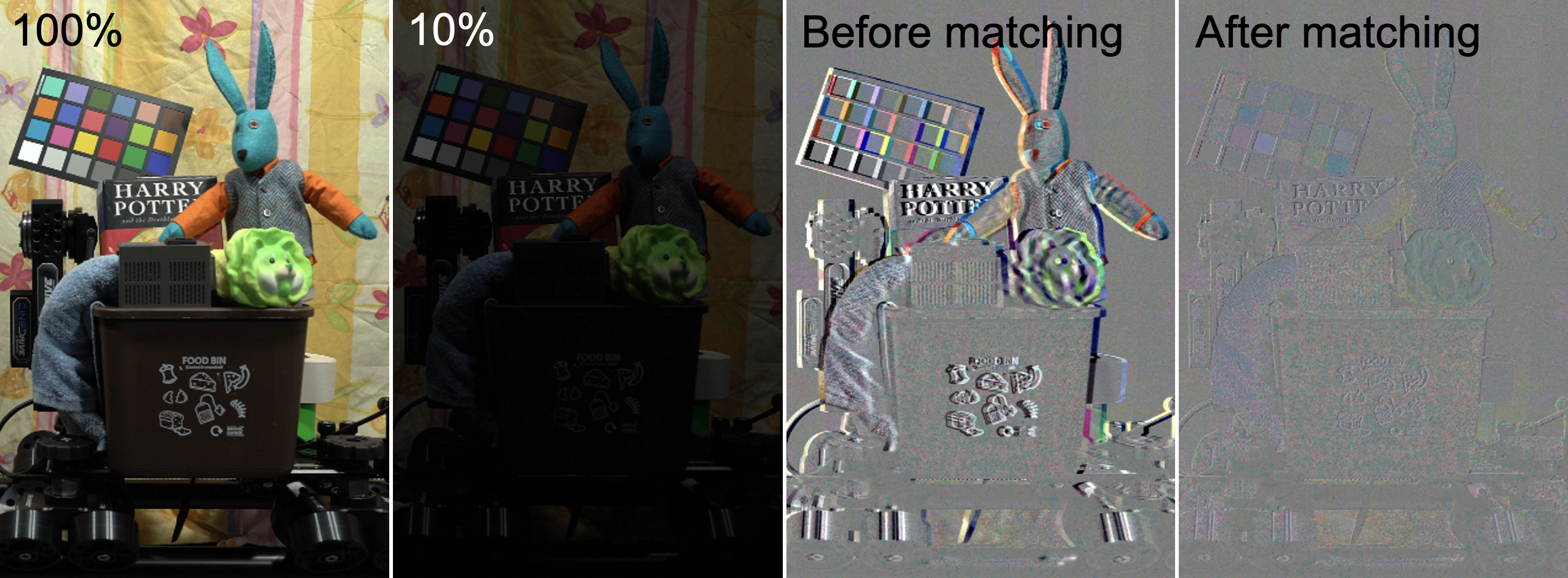}
  \caption{Moving bunny scene with static background showing pixel value difference between the normal-light frame and the adjusted low-light frame before and after frame matching. Gray indicates zero error.}
  \label{fig:bunny}
\end{figure}


\subsection{The dataset}
\label{ssec:dataset}
\begin{figure*}[t]
    \centering
    \includegraphics[width=\textwidth]{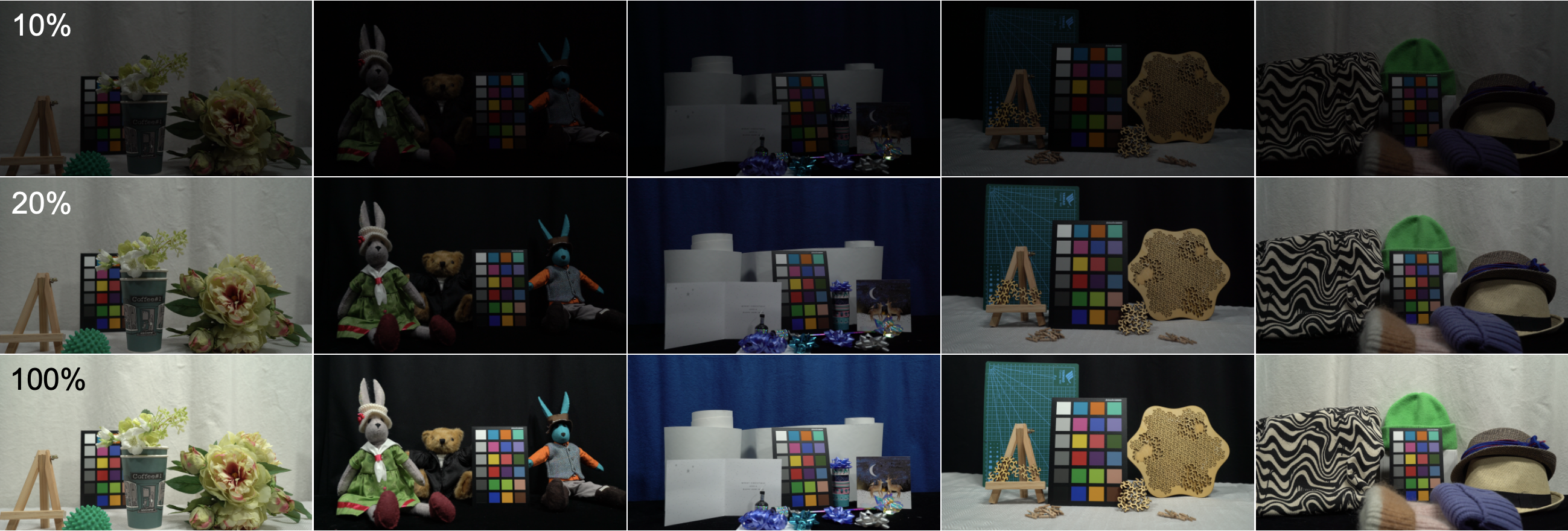}
    \caption{The proposed dataset: (From left to right) flowers, soft toy, wrapping, wood board, hats. (From top to bottom) Light levels of 10\%, 20\%, and normal light (100\%).}
    \label{fig:examplescenes}
\end{figure*}

In total, we acquired 40 scenes of two different low light levels, including 6 scenes of static background. This results in 31,800 image pairs with various motion settings, as detailed in Table \ref{tab:scenedescription}. Example scenes are presented in Fig. \ref{fig:examplescenes}, where 100\% indicates normal light which can be used as reference or ground truth. The number of frames varied based on the dolly's speed.

We captured both static backgrounds with moving objects and complete dynamic scenes. For the static background scenes, we placed the objects on the dolly, and the camera on the tripod. In contrast, for creating more dynamic content, we mounted the camera on the dolly, creating in both linear and nonlinear motions.  Three different camera positions and movements relative to the dolly's direction were configured, denoted as follows in Table \ref{tab:scenedescription}: i) \textit{parallel}, where the camera axis remained perpendicular to the dolly's movement direction, ii) \textit{angle}, where the camera axis was positioned at a 30-degree angle to the dolly's movement direction (as shown in Fig. \ref{fig:cinedrive}), and iii) \textit{rotation}, indicating camera rotation in its yaw angle while the dolly was in motion. Dolly speeds were set to be constant or varied after clipping, denoted as \textit{linear} and \textit{nonlinear}. Fig. \ref{fig:motions} displays various motions captured in the videos. The x-t planes of three example videos reveal different movements. Mario3 exhibits linear motion of the foreground object with a static background. In the dynamic scene of Book1, there is a transition from fast motion at the beginning, slowing down in the middle, and speeding up towards the end of the video. In the Mario1 scene, there is nonlinear movement from left to right and then a return to the left of the scene.

\begin{figure}[t]
    \centering
    \includegraphics[width=0.75\columnwidth]{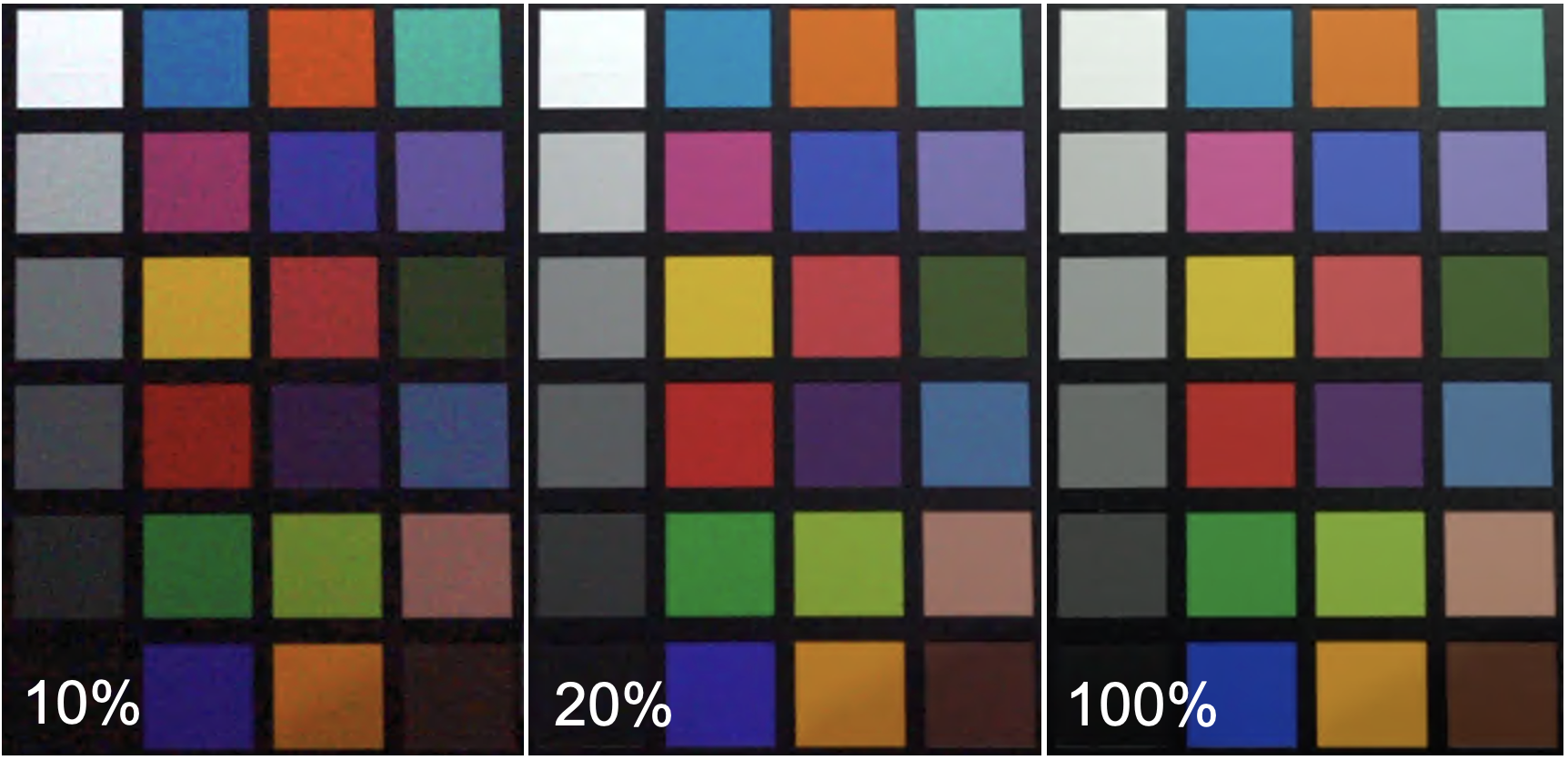}
    \caption{Cropped images (lego and kitchen scenes at 350$\times$350 and 140$\times$140 pixels, respectively) with histogram matching to the reference (normal light) to visualise noise at different levels of light. (From left to right) Light levels of 10\%, 20\%, and normal light (100\%).}
    \label{fig:crop_histadjust}
\end{figure}

Fig. \ref{fig:crop_histadjust} reveals the noise present in the low-light scenes (for visualization, we applied histogram matching to ground truth). It's worth noting that color correction is not needed extensively, as we have included the standard color palette in the scene, which should also assist in the training process. 

\section{PCDUNet framework}

\begin{figure}[t]
    \centering
    \includegraphics[width=0.75\columnwidth]{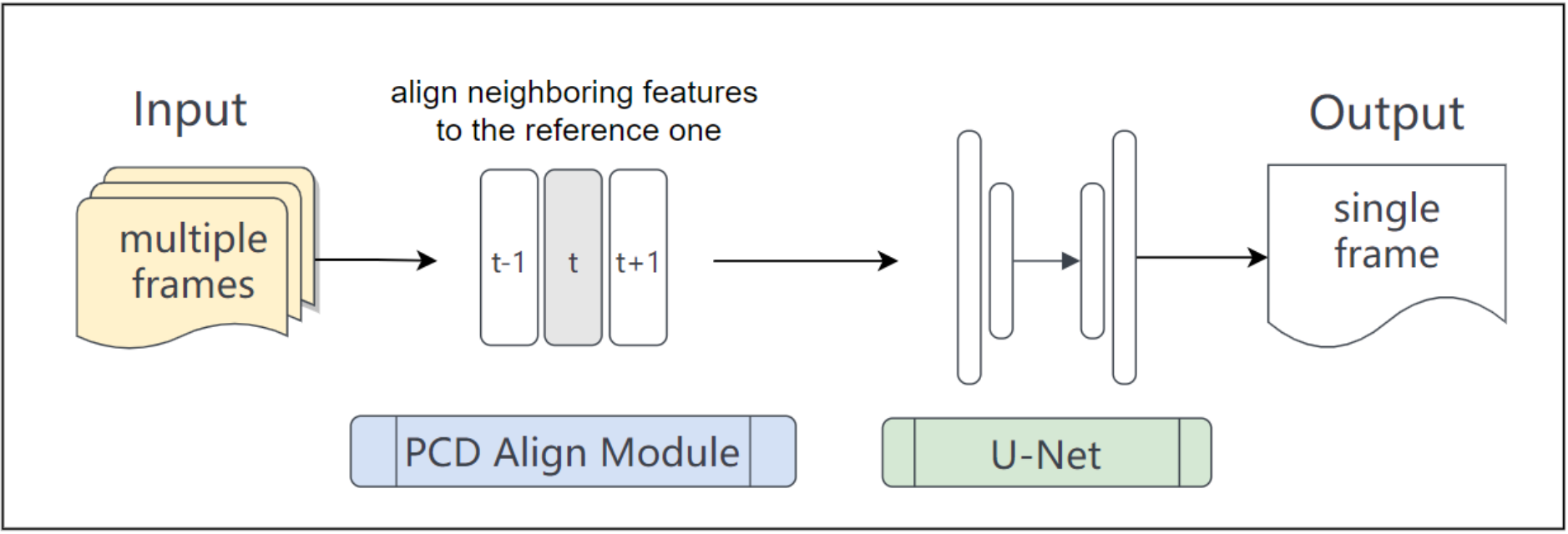}
    \caption{PCDUNet framework}
    \label{fig:PCDUNet}
\end{figure}

We analysed and tested our dataset with PCDUNet framework. We employ a Pyramid, Cascading and Deformable (PCD) alignment module as used in EDVR \cite{Wang:EDVR:2019}, followed by UNet, as shown in Fig. \ref{fig:PCDUNet}.

\section{Experiments}
\label{sec:experiments}

This section examines the importance of fully registered video pairs in the development of low-light video enhancement methods. We analyze distortions in low-light video sequences and compare our proposed datasets with existing ones. We achieve this by using them to train several state-of-the-art methods for low-light enhancement and video restoration.
In all experiments that used our dataset, we randomly selected 32 scenes for training and allocated the remaining scenes for testing. These training and testing sets were fixed to all experiments.

\subsection{Importance of fully registered data}
\label{ssec:fullyregis}

The training process of supervised learning optimizes the model by minimizing loss. The most commonly used loss functions in image and video enhancement are $\ell_1$, $\ell_2$, charbonnier, SSIM, and VGG loss \cite{anantrasirichai:AI:2022}. While $\ell_1$ and $\ell_2$ require pixel-wise alignment between the predicted output and the ground truth, SSIM and VGG losses allow for a slight shift between them.

We demonstrate the importance of a fully registered dataset for supervised learning approaches in image enhancement by retraining three network architectures: UNet \cite{ronneberger2015unet}, Conditional GAN \cite{Wang:hight:2018}, and Transformer (SwinIR \cite{Liang:SwinIR:2021}) with our dataset before and after frame synchronization, as explained in Section \ref{ssec:gtgen}. Prior to synchronization, the maximum misalignment is 20 pixels (considering that the frame resolution is 1,920$\times$1,080 pixels). We also conducted experiments with synthetic misalignments of 3, 5, and 10 pixels between the input and ground truth and the results are reported in supplementary material. 

\begin{figure}[t]
  \centering
\includegraphics[width=0.8\linewidth]{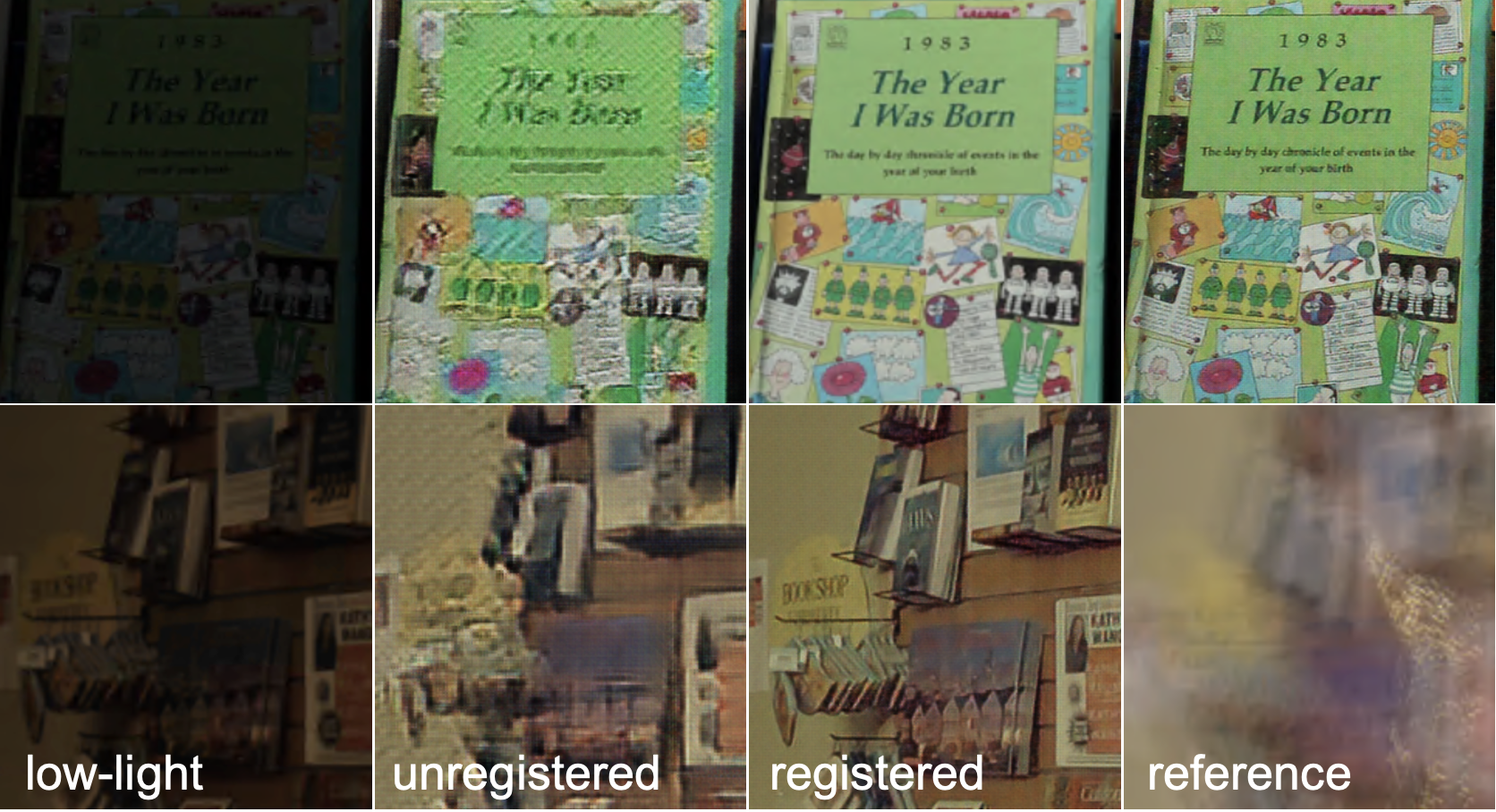}
   \caption{Comparison of training CGAN \cite{Wang:hight:2018} with the unregistered and registered datasets. Top and bottom rows are the cropped testing dynamic scenes of our dataset and DRV dataset, respectively.}
   \label{fig:studyregister}
\end{figure}

Table \ref{tab:studyregister} shows the performance impact of low-light enhancement when the models are trained with either non-registered or registered datasets. As expected, performance decreases when trained with unregistered datasets, with a noticeable effect when using the $\ell_1$ loss. Fig. \ref{fig:studyregister} shows the results of testing dynamic scenes in our dataset and the DRV dataset, both trained with our dataset before and after alignment (the top row used SSIM loss, while the bottom row used $\ell_1$ loss). The results clearly exhibit artifacts, which occur because the network cannot align the predicted output with the ground truth to minimize the loss. The DRV dataset does not provide ground truth but offers a long-exposure version for brightness reference.

\begin{table}[t]
    \centering
    \caption{Comparison of results (PSNR/SSIM) using non-registered (No) and registered (Yes) datasets with different loss functions for training.}
    \small
    \begin{tabular}{c|cc|cc|cc}
    \toprule
        loss & \multicolumn{2}{|c|}{$\ell_1$} & \multicolumn{2}{|c|}{SSIM} & \multicolumn{2}{|c}{VGG} \\ \hline
        Register? & No & Yes & No & Yes & No & Yes  \\ 
        \hline
         \multicolumn{7}{c}{PSNR} \\
         \hline
        UNet & 20.05 & 20.72 & 19.45 & 20.37 & 19.89 & 20.94\\
        CGAN & 20.75  & 26.60 & 23.91 & 26.14 & 24.60 & 28.41\\
        SwinIR & 20.69 & 23.62 & 18.45 & 22.04 & - & - \\
        \hline
         \multicolumn{7}{c}{SSIM} \\
         \hline
        UNet & 0.760 & 0.761 & 0.731 & 0.785 & 0.752 & 0.753 \\
        CGAN & 0.702 & 0.862 & 0.786 & 0.861 & 0.796 & 0.911\\
        SwinIR & 0.693 & 0.745 & 0.677 & 0.732 & - & -\\
    \bottomrule
    \end{tabular}
    \label{tab:studyregister}
\end{table}

\subsection{Influence of different light levels in datasets}

Our dataset provides two light levels in low-light environments. This allows us to investigate whether the improved video quality depends on the light levels present in the training data. We trained our PCDUNet using data from only the 10\% lighting, 20\% lighting, and both light levels. Then we tested them with the same and cross light levels. The results are shown in \ref{tab:studylightlevels}. It is obvious that the model trained with the same lighting level as the testing data performs better than one trained with different lighting levels. However, if the light level is unknown, the model trained with various light levels exhibits more robustness, as shown in the third column of the table.

\begin{table}[t]

    \centering
     \caption{Performance of low-light video enhancement method, PCDUNet, when training with different light levels}
     \small
    \begin{tabular}{c|ccc}
    \toprule
    test & 10\% & 20\% & 10\%+20\%  \\ \hline
        train & PSNR/SSIM & PSNR/SSIM & PSNR/SSIM  \\ 
        \hline
        10\% & 29.52/0.894 & 12.76/0.669 & 21.33/0.784\\
        20\% & 13.49/0.620 & 31.36/0.936 & 22.24/0.775\\
        10\%+20\% & 17.81/0.728 & 31.17/0.932 & 24.52/0.845\\
    \bottomrule
    \end{tabular}
   
    \label{tab:studylightlevels}
\end{table}

\subsection{Dataset comparison}

\begin{figure}[t]
  \centering
   \includegraphics[width=\linewidth]{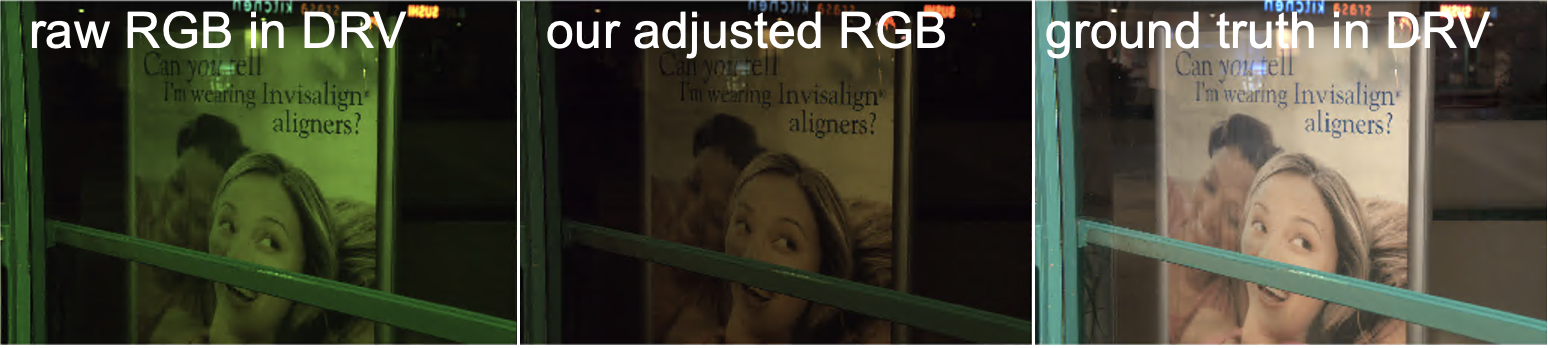}
   \caption{DRV dataset \cite{Chen_2019_ICCV} and our green channel adjustment}
   \label{fig:DRVdata}
\end{figure}

We evaluated the training effectiveness of our datasets with low-light video enhancement methods, including SMOID \cite{Jiang:learn:2019}, SMID \cite{Chen_2019_ICCV}, SDSD \cite{wang2021sdsd}, as well as state-of-the-art video restoration techniques, such as EDVR \cite{Wang:EDVR:2019}, VRT \cite{liang2022vrt}, and our PCDUNet, as utilized in a previous study. We compared the performance of these models trained with our datasets to those trained with existing datasets, namely DRV \cite{Chen_2019_ICCV} and SDSD \cite{wang2021sdsd}. The RGB data of DRV\footnote{the data available online are in standard RGB format with 8-bit depth} appears excessively green, which is not natural (see Fig. \ref{fig:DRVdata}). Upon observation, we noticed that the values of the green channel are approximately twice as high as those in the other color channels. Consequently, we adjusted the green channel by reducing its values by half to ensure a fair comparison when using the networks that accept standard RBG inputs.

When dividing the data for training and evaluation, we followed to the parameters outlined in the original papers. The training-to-testing ratios were 73:27 for DRV and 84:16 for SDSD. In our dataset, we chose an intermediary approach by randomly splitting our datasets into 80\% for training and 20\% for testing. The training parameters were also aligned with those outlined in the original papers, ensuring consistency across all datasets for a fair comparison.
The results, presented in Table \ref{tab:alltest}, are the average performance across three datasets. 

\begin{table}[t]
    \centering
    \caption{Performance comparison of the models trained with different datasets. The objective results shown here are the average of three testing datasets.}
    \small
    \begin{tabular}{c|ccc}
    \toprule
        Dataset & DRV \cite{Chen_2019_ICCV} & SDSD \cite{wang2021sdsd} & {Our dataset}  \\ \hline
        Methods & PSNR/SSIM & PSNR/SSIM & PSNR/SSIM \\
        \hline
        \multicolumn{4}{c}{Without postprocessing} \\
        \hline
        UNet  \cite{ronneberger2015unet} & 13.73/0.328 & 15.89/0.617 & 19.44/0.710\\
        CGAN \cite{Wang:hight:2018} & 17.53/0.490 & 15.52/0.461 & 19.80/0.621  \\
        RIDNet \cite{Anwar:Real:2019} & 15.30/0.544 & 15.02/0.498 & 20.48/0.709\\ 
        SwinIR \cite{Liang:SwinIR:2021} & 17.07/0.696 & 12.11/0.484 & 20.06/0.726\\
        SDSD \cite{wang2021sdsd} &  15.29/0.542 &  15.01/0.496 & 16.03/0.616 \\
        PCDUNet & 16.67/0.512 & 16.11/0.636 & 20.80/0.757\\
        \hline
        \multicolumn{4}{c}{Postprocessing with histogram matching to a reference} \\
        \hline
        CGAN \cite{Wang:hight:2018} & 29.67/0.836 & 28.26/0.800 & 32.91/0.890  \\
        SwinIR \cite{Liang:SwinIR:2021} & 28.31/0.839 & 26.68/0.789 & 30.86/0.908\\
    \bottomrule
    \end{tabular}
    
    \label{tab:alltest}
\end{table}

\subsection{Dataset comparison when lighting reference is available}

As brightness perception in videos can be subjective, we conducted evaluations considering a preferable brightness level as postprocessing. To do this, we used the first frame from the ground truth sequence as the reference. We performed histogram matching between this reference frame and the first frame of the enhanced output. Subsequently, each subsequent enhanced frame was adjusted concerning its previous frame, ensuring consistency throughout. Table \ref{tab:postprocessing} shows the results comparing postprocessing against scenarios with no brightness reference. Interestingly, the PSNR and SSIM of the testing SDSD dataset exhibit the most significant improvement across all models trained with various datasets, with more than threefold increase. This suggests that, when disregarding brightness considerations, enhancing the SDSD dataset is notably more straightforward. Primarily, this trend arises due to the SDSD dataset containing a smaller amount of noise compared to other datasets. Moreover, its ground truth brightness significantly differs from the normal light defined in other datasets, resulting the models perform worse when trained with the SDSD dataset. Comparing training and testing on the same dataset, the postprocessing improves the performances by approximately 25\%, 200\%, and 5\% (average of PSNR and SSIM) when using DRV, SDSD and our dataset, respectively. This implies that our dataset has better consistent brightness of the ground truth.

\begin{table}[t]
    \centering
    \caption{Performance comparison of CGAN \cite{Wang:hight:2018} when trained on different datasets, reporting on individual testing datasets with and without postprocessing.}
    \small
    \begin{tabular}{c|ccc}
    \toprule
       & \multicolumn{3}{c}{Test on} \\ \hline
        Train & DRV \cite{Chen_2019_ICCV} & SDSD \cite{wang2021sdsd} & {Our dataset}  \\ 
        \hline
        \multicolumn{4}{c}{Without postprocessing} \\
        \hline
        DRV \cite{Chen_2019_ICCV}  &	21.25/0.657 &	11.86/0.254  & 19.48/0.559 \\
        SDSD \cite{wang2021sdsd}  &	17.76/0.518 &	9.56/0.307  & 19.26/0.560\\
        {Our dataset} & 18.54/0.611 & 12.45/0.340 & 28.41/0.911 \\
        \hline
        \multicolumn{4}{c}{Postprocessing with histogram matching to a reference} \\
        \hline
        DRV \cite{Chen_2019_ICCV} 	& 27.33/0.810 &	36.01/0.904 & 25.67/0.794\\
        SDSD \cite{wang2021sdsd}  &	23.57/0.715 &	37.32/0.912  & 23.89/0.765 \\
        {Our dataset} & 26.67/0.812 &	40.32/0.933 & 31.78/0.925  \\
    \bottomrule
    \end{tabular}
    
    \label{tab:postprocessing}
\end{table}

\section{Conclusion}
\label{sec:conclude}
This paper introduces a novel dataset for low-light video enhancement. Comprising 40 dynamic scenes captured under two low-light levels, our dataset ensures full registration in both spatial and temporal dimensions with the corresponding normal-light videos which can be used as ground truth. This benefits the development of supervised learning models and supports  full-reference quality assessment. We illustrate the effectiveness of our dataset through a comprehensive study and experimental comparisons with existing low-light video datasets.


\begin{table}
	\centering
  \caption{Scene descriptions: Frame denotes the total number of frames. Speed represents the approximate dolly speed. Camera describes the camera's movement relative to the dolly, with `parallel' and `angle' denoting fixed angles, and `rotation' indicating camera rotation in its yaw angle. Motion uses `linear' and `nonlinear' to distinguish constant and varied speeds, respectively. `(S)' denotes static background.}
	\footnotesize
		\begin{tabular}{rlcccc}
		\toprule
         \# &  Name &  Frames & Speed & Camera & Motion\\
	    \midrule
        	1. & color chart & 270 & slow & parallel & linear \\
        	2. & Animals1 & 425 & slow & angle & linear \\
        	3. & Animals2 & 600 & slow & rotation & nonlinear\\
                4. & color sticks & 350 & slow & parallel & linear \\
                5. & Bunnies & 300 & fast & parallel & nonlinear \\
                6. & Lego & 214 & fast & parallel & linear \\
                7. & Hats & 650 & slow & parallel & nonlinear \\
                8. & Soft toys & 400 & slow & parallel & nonlinear \\
                9. & Kitchen & 257 & fast & rotation & nonlinear\\
                10. & Messy toys & 210 & fast & parallel & nonlinear \\
                11. & Gift wrap & 260 & fast & parallel & nonlinear \\
                12. & Woods & 240 & fast & angle & linear\\
                13. & Wires & 240 & fast & angle & linear\\
                14. & color balls & 240 & fast & parallel & linear \\
                15. & Flowers & 330 & mix & rotation & nonlinear \\
                16. & Faces1 & 730 & slow & parallel & nonlinear \\
                17. & Faces2 & 400 & fast & parallel & linear \\
                18. & Books1 & 500 & fast & parallel & linear \\
                19. & Books2 & 554 & fast & parallel & nonlinear \\
                20. & Books3 & 380 & fast & parallel & linear \\
                21. & Mario1 & 700 & fast & parallel & nonlinear \\
                22. & Mario2 & 340 & fast & angle & linear \\
                23. & Messy wires & 580 & slow & angle & nonlinear\\
                24. & Figures1 & 450 & slow & parallel & nonlinear \\
                25. & Figures2 & 200 & fast & rotation & nonlinear\\
                26. & Cameras & 300 & fast & rotation & linear \\
                27. & Cloths1 & 430 & fast & rotation & nonlinear \\
                28. & Cloths2 & 330 & fast & mix & nonlinear \\
                29. & Mix1 & 600 & mix & parallel & nonlinear \\
                30. & Mix2 & 550 & slow & angle & linear \\
                31. & Mix3 & 300 & slow &  parallel & linear\\
                32. & Mix4 & 450 & mix & parallel & nonlinear\\
                33. & Mix5 & 200 & fast & rotation & linear \\
                34. & Mix6 & 250 & fast & rotation & linear \\
                35. & Bunny (S) & 480 & slow & parallel & linear \\
                36. & Eeyore (S) & 480 & slow & parallel & linear \\
                37. & Mario3 (S) & 370 & slow & parallel & nonlinear \\
                38. & Mario4 (S) & 280 & fast & parallel & nonlinear \\
                39. & Books4 (S) & 400 & slow & parallel & nonlinear \\
                40. & Books5 (S) & 660 & mix & parallel & nonlinear \\
        \bottomrule
		\end{tabular}
	\label{tab:scenedescription}
 
\end{table}


%








\bibliographystyle{IEEEtran}
\bibliography{IEEEabrv,bibtex/bib/lowlight}
%

%








\end{document}